# A survey of methods to ease the development of highly multilingual text mining applications

Ralf Steinberger

*European Commission – Joint Research Centre (JRC)*

*Via Fermi 2749, 21027 Ispra (VA), Italy*

E-mail: Ralf.Steinberger@jrc.ec.europa.eu

URL: http://langtech.jrc.ec.europa.eu/RS.html

**Abstract.** Multilingual text processing is useful because the information content found in different languages is complementary, both regarding facts and opinions. While Information Extraction and other text mining software can, in principle, be developed for many languages, most text analysis tools have only been applied to small sets of languages because the development effort per language is large. Self-training tools obviously alleviate the problem, but even the effort of providing training data and of manually tuning the results is usually considerable. In this paper, we gather insights by various multilingual system developers on how to minimise the effort of developing natural language processing applications for many languages. We also explain the main guidelines underlying our own effort to develop complex text mining software for tens of languages. While these guidelines – most of all: extreme simplicity – can be very restrictive and limiting, we believe to have shown the feasibility of the approach through the development of the *Europe Media Monitor* (EMM) family of applications (http://press.jrc.it/overview.html). EMM is a set of complex media monitoring tools that process and analyse up to 100,000 online news articles per day in between twenty and fifty languages. We will also touch upon the kind of language resources that would make it easier for all to develop highly multilingual text mining applications. We will argue that – to achieve this – the most needed resources would be freely available, simple, parallel and uniform multilingual dictionaries, corpora and software tools.

Abbreviations: ML: Machine Learning; NER: Named Entity Recognition; JRC: Joint Research Centre; EC: European Commission; EU: European Union; MT: Machine Translation; EMM: Europe Media Monitor; LREC: Language Resources and Evaluation Conference;

MeSH: Medical Subject Headings; TAC: Text Analysis Conference; LDC: Linguistic Data Consortium; CoNLL: Conference on Computational Natural Language Learning; ELDA: Evaluations and Language Resources Distribution Agency; GATE: General Architecture for Text Engineering;

## 1. Introduction

The share of non-English documents on the internet is rising continuously. While many private users will only be interested in finding monolingual information in their own language, the need for multilingual information retrieval, information extraction and cross-lingual information access for professionals, organisations and businesses is rising steadily. Starting from the premise that we need multilingual text mining tools, the question we would like to ask here is: *How can we avoid that the development of (any) text mining application for N languages takes N times the effort of developing them for one language*. It is generally acknowledged that developers benefit from the experience of having produced tools in one or more languages before, and that the existence of an efficient implementation infrastructure is extremely important (e.g. Maynard et al. 2002). Such software building blocks can include, for instance, a grammar implementation formalism, tools for marking up text, debugging tools, automatic evaluation tools and procedures, etc. Furthermore, simple applications like sentence splitters are typically so similar for different languages that – once one exists – the same tool is usually quickly adapted to new languages. We will thus try to take the effort of developing the infrastructure out of the equation. The question should thus be reformulated: *Assuming that you have already developed text mining applications for some languages, how can you limit the effort involved in the development of such applications for several other languages*.

In the next section, we will try to demonstrate the need for multilingual text processing and to show that most application providers offer monolingual tools or tools covering a few commonly spoken languages. In Section (3), we will describe the type of data we work with (mostly news) and give a short overview of the functionality of the *Europe Media Monitor* family of applications. In Section (4), we will then try to answer the main question asked here. First, we will summarise insights by other multilingual system developers (0) and discuss the contribution of Machine Learning methods (0) – in our view an extremely promising approach to go highly multilingual. We will then present our own guidelines on how to minimise the effort of multilingual tool development (0), which – of course – largely overlap with those proposed by others. In Section 5, we will give some examples of what these insights and guidelines concretely mean for the development of a small selection of natural language processing tools. One obvious bottleneck for the development of multilingual tools is the lack of linguistic resources. In Section 6, we thus share our view on which kind of resources would be particularly beneficial to achieve highly multilingual text mining applications. Section 7 summarises and concludes.

## 2. Motivation for multilingual text mining

The *Joint Research Centre* (JRC) is the scientific-technical arm of the European Commission (EC). The European Union (EU) institution EC is a multinational organisation with strong links also to countries outside the EU. It is thus natural that multilinguality plays a big role inside the organisation. However, experience with the many partners and customers of the JRC shows clearly that even many national organisations have a need for highly multilingual text processing applications.

The JRC receives frequent requests to monitor media reports in dozens of languages, involving news gathering, classification, information extraction and analysis. The JRC's users consist of EU institutions, state organisations inside its 27 Member States, institutions of partners outside the EU (e.g. in the USA, Canada, China, etc.), as well as international organisations (including various United Nations and pan-African sub-organisations). These users have a wide range of interests so that not only media reports in the 23 official EU languages need to be monitored, but also, for instance, those in the languages of the EU's neighbouring countries, of the world's crisis areas and of political partner countries around the world.

To give a concrete example: Public Health organisations around the world monitor any threats to the populations of their counties – be they chemical, biological, radiological or nuclear (CBRN). For that purpose, they not only gather information on communicable diseases, etc. from their hospitals (*indicator-based* risk monitoring), but they also scan online news articles and government websites to find out about the outbreak of communicable diseases, etc. (*event-based* risk monitoring; Linge et al. 2009). In the era of high mobility and mass long-distance travel, the risk of contracting a disease (e.g. the human influenza virus, also referred to as 'swine flu' and H1N1), taking it home and passing it on to others is so big that the Public Health community follows the situation around major tourist destinations and locations for international religious and sports-related mass gatherings thoroughly, by monitoring international media reports published around the world.

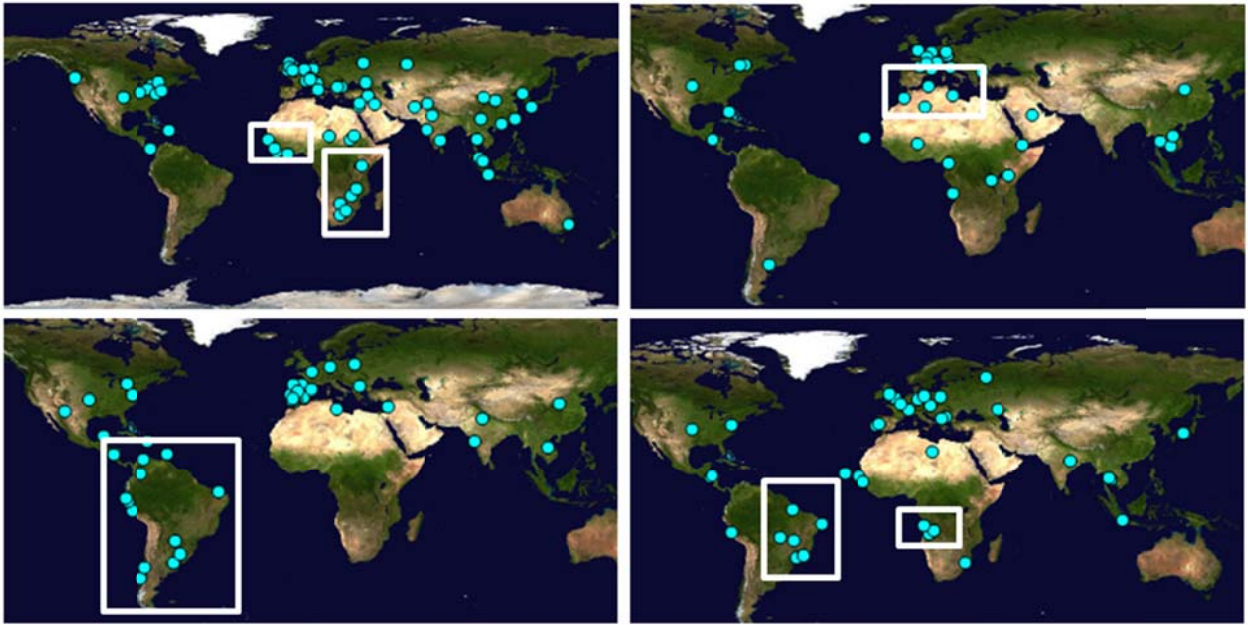

**Figure 1.** The four maps show the complementary locations mentioned in health-related news published in the same time window in the four world languages English, French, Spanish and Portuguese (from top left to bottom right).

It is our experience that *multilingual* media monitoring is not only a luxury, but – due to the *information complementarity* in the news across different languages – an urgent requirement. Large events and events that are in the focus of the world media (e.g. reports from conflict areas such as Iraq or Israel, or reports about human bird flu cases) will usually be translated into English and other world languages. However, many smaller events rarely make it into the international news, including local reports on the outbreak of more common diseases (e.g. tuberculosis or malaria), or reports about pastoral conflicts in Africa, although this type of report may be important to organisations monitoring Public Health or country stability. **Figure 1** gives a good indication of cross-lingual information complementarity occurring in targeted real-life news.

Information complementarity not only applies to contents, but also to opinions: by considering points of view from around the world, readers will get a *less biased* and more balanced view on world events. To give only one simple example: Daily and long-term social network analysis across various countries and languages (Pouliquen et al. 2007a) has shown that the most central personalities are usually the respective leaders of state. When only reading English language news, readers will thus get an inflated impression of the importance of the US President and the British Prime Minister, while the readers of Russian, Arabic or Spanish language news will get quite a different impression.

The most common approach to capturing information published in foreign languages is the use of Machine Translation into one target language (e.g. English) and to apply information filtering and extraction tools in that target language. A limitation of this approach is that proper names and specialist terms are frequently badly translated so that information can easily get lost. Our own insight (supported by the *native language hypothesis* observed by Larkey et al. 2004) is that information filtering in the source language is more efficient than filtering machine-translated text. In the USA, Machine Translation is nevertheless an attractive solution, as there is only one official national language. However, when looking at Europe, Asia and other parts of the world, it becomes clear that the situation in the US is an exception rather than the rule, as there is no agreement on one common language.

News aggregators such as Google News[1], Yahoo News[2] and EMM[3] already gather and cluster news in many languages (currently 46, 32 and 50 languages, respectively – status February 2011), but most of the more complex systems carrying out some level of analysis of the gathered texts are monolingual, including *SiloBreaker*[4], *NewsVine*[5] and *DayLife*[6]. The news analysis systems *NewsTin*[7] and the EMM product *NewsExplorer*[8] are notable exceptions, covering 11 and 20 languages, respectively.

We believe that the main reason for the existence of monolingual analysis systems is the large effort required to produce text processing software for new languages. In the worst case, the effort required to develop tools in N languages is N times the effort of developing monolingual software, but various multilingual system developers have found methods to minimise this effort. These insights will be the main focus of the rest of the paper.

---

[1] See http://news.google.com . All websites mentioned here were last visited in the week of 15 February 2011.
[2] See http://news.yahoo.com/ .
[3] See http://emm.newsbrief.eu/ .
[4] See http://www.silobreaker.com/ .
[5] See http://www.newsvine.com/ .
[6] See http://www.daylife.com/ .
[7] See http://www.newstin.com/ .
[8] See http://emm.newsexplorer.eu/. NewsExplorer processes news articles in Arabic, Bulgarian, Danish, Dutch, English, Estonian, Farsi, French, German, Italian, Norwegian, Polish, Portuguese, Romanian, Russian, Slovene, Spanish, Swahili, Swedish and Turkish.

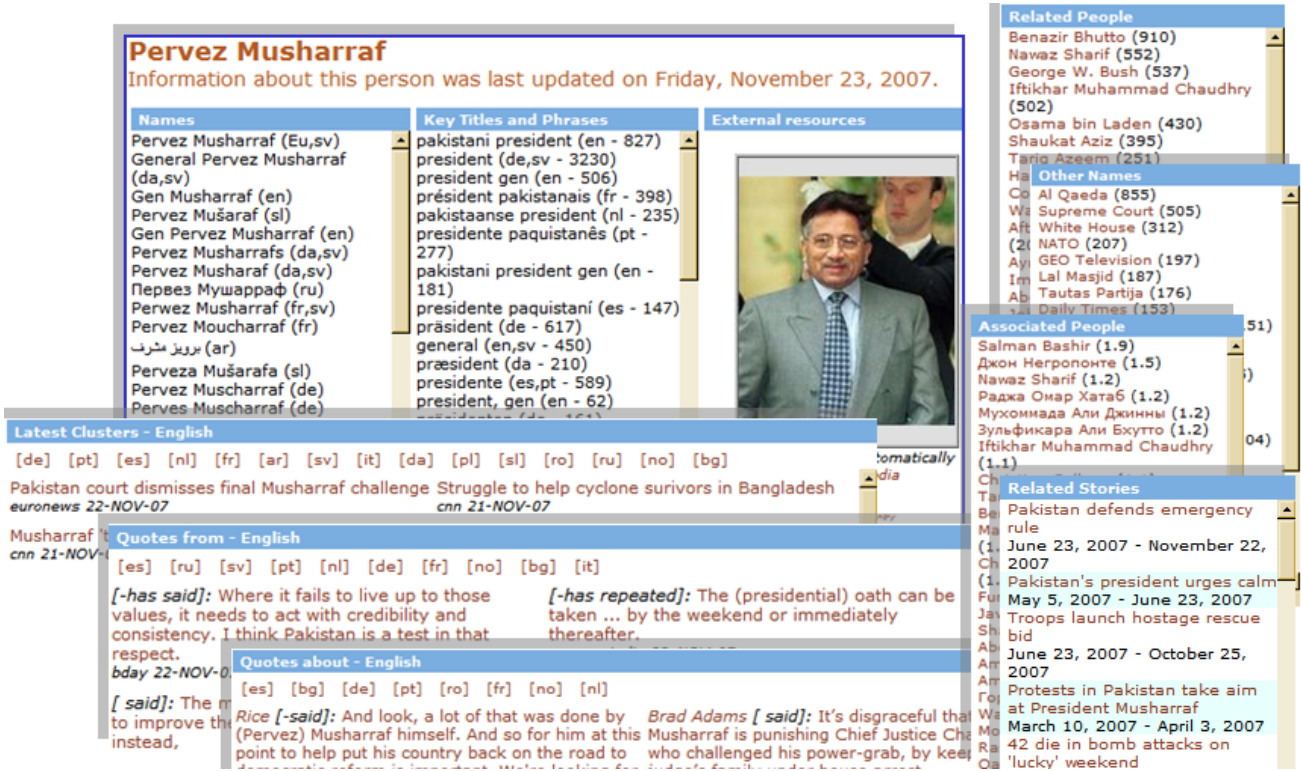

**Figure 2.** Named entity-related information extracted and aggregated by the EMM application *NewsExplorer* from news in 20 languages, including: name variants, titles, latest clusters and 'stories', quotes by and about that person, ranked lists of persons and other entities mentioned historically in the same clusters.

## 3. The Europe Media Monitor family of applications

The *Europe Media Monitor* (EMM, Steinberger et al. 2009) is the basic engine that gathers an average of about 100,000 news articles per day in approximately 50 languages[9], from about 2,500 hand-selected web news sources, from a couple of hundred specialist and government websites, as well as from about twenty commercial news providers. EMM visits the news web sites up to every five minutes to search for the latest articles. When news sites offer RSS feeds, EMM makes use of these, otherwise it extracts the news text from the often complex HTML pages. All news items are converted to Unicode. They are processed in a pipeline structure, where each module adds additional information. Whenever files are written, the system uses UTF-8-encoded RSS format.

---

[9] As of February 2011, the website actually lists 54 languages, but some of them are extremely low-volume and EMM may not capture news in these languages every day.

The EMM news gathering engine feeds its articles into the four fully-automatic public news analysis systems (accessible via http://emm.jrc.it/overview.html), and to their non-public sister applications (Steinberger et al. 2009). The major concern of *NewsBrief* and *MedISys* is breaking news and short-term trend detection (topic tracking), early alerting and up-to-date category-specific news display. *NewsExplorer* focuses on daily overviews, long-term trends (topic tracking), linking of related news across languages, in-depth analysis and extraction of information about people and organisations (see Figure 2). *EMM-Labs* is a collection of more recent developments, including a multilingual event scenario template filling application, a multilingual multi-document summarisation demonstrator, and various tools to visualise extracted news data. For *NewsBrief* and *MedISys*, there are different access levels, distinguishing the entirely public web sites from an EC-internal website. The public websites do not contain commercial sources and offer slightly reduced functionality.

The following JRC-developed text mining methods and tools are used and closely integrated in EMM; if not mentioned otherwise, they work for 20 languages: document clustering and Boolean classification (50 languages); breaking news detection and automatic user notification (50 languages); Named Entity Recognition (persons, organisations); name variant matching (i.e. string distance calculation, including across scripts); geo-tagging (recognition, disambiguation and grounding for map-display); quotation recognition (reported speech by and about named entities); multi-label classification using the thousands of categories from the Eurovoc[10] thesaurus; multi-monolingual topic tracking (to detect 'stories') and aggregation of information per 'story'; cross-lingual news cluster linking (available for the majority of the 190 possible language pairs); social network generation based on information extracted from multilingual news (based on co-occurrence, and also on who mentions whom in reported speech); detailed scenario template filling for events causing victims (violence, natural disasters, accidents, disease outbreaks, etc.; seven languages); visualisation (using geographical maps, trends, social networks, etc.).

EMM was mostly developed to serve the interests of the European Institutions and their international partners, but the public web pages are also visited by an average of 30,000 anonymous users per day.

---

[10] See http://europa.eu/eurovoc/. Automatic Eurovoc indexing has been trained for 22 EU languages.

## 4. How to achieve multilinguality

Many individual natural language processing applications have been developed for several languages, covering varying numbers of languages. We have not found many publications directly addressing the issue on how to minimise the effort of multilingual tool development, but several that describe the efforts of adapting a certain tool to a new language. Typically, these applications are named entity recognition systems or syntactic parsers. Section 0 contains a list of ideas found in such publications. Section 0 addresses the role of Machine Learning approaches, which seem to be particularly useful to achieve multilinguality. Section **4.3** briefly highlights the use of methods for cross-lingual projection. Section **0** summarises our own approach which, obviously, in many cases, overlaps with that of other developers.

**Related work: Insights by other multilingual developers**

Multiple authors have described work on developing resources and tools for a number of different languages. This was typically done by reusing the resources from a first language and adapting them to new languages (e.g. Gamon et al 1997; Rayner & Bouillon 1996; Pastra et al. 2002; Carenini et al. 2007; Maynard et al. 2003). Practical tips from various system developers for achieving multilinguality include the use of **Unicode** and of the usage of **virtual keyboards** to enter foreign language script (Maynard et al. 2002); **modularity** (Pastra et al. 2002; Maynard et al. 2002); **simplicity** of rules and the lexicon (Carenini et al. 2007; Vergne 2002); **uniform input and output structures** (Carenini et al. 2007; Bering et al. 2003); and the use of **shared token classes** that are ideally based on surface-oriented features such as case, hyphenation, and includes-number (Bering et al. 2003). SProUT grammar developers took the interesting approach of using shared resources between languages (lexica, gazetteers, grammar rules) for named entity recognition in seven languages, and of splitting the multilingual grammar rule files (Bering et al. 2003): some files contain rules that are applicable to several languages (e.g. to recognise dates of the format *20.10.2010*) while others contain language-specific rules (e.g. to cover *20th of October 2010*). The fact that this latter date format, and others, can also be captured by using language-independent patterns was shown by Ignat et al. (2003).

Both Maynard et al. (2002) and Pastra et al. (2002) point out that the usage of **theory-neutral data types** is an advantage for the Language Engineering architecture GATE because it facilitates reuse. This does make sense for a platform that is meant to be used by many groups for many purposes. However, there are several grammar developers who point out that adhering to **grammar theories** is very efficient because they separate universal rules from language-specific parameters and differences. For instance, Bender & Flickinger (2005) highlight the benefits of adhering to Head-Driven Phrase Structure Grammar (HPSG) for writing multilingual general-purpose grammars. They even propose to *generate* starter grammars for new languages automatically, based on a number of linguistic features of that language. Gamon et al. (1997) report that the framework of Universal Grammar allows them to create a generic grammar that "can easily be parameterised to handle many languages". Interestingly, they provide detailed information on the percentage of grammar rule overlap between their original English general-purpose Microsoft-NLP grammar and the German, French and Spanish grammars they derived from the English version. Wehrli (2007), using Chomsky's generative grammar to build parsers for six languages, stipulates that the design he adopts "makes it possible to 'plug' an additional language without any change or any recompilation of the system. It is sufficient to add the language-specific modules and lexical databases". Ranta (2009, e.g. pp. 47ff), having worked within the Grammatical Framework on fourteen languages, also addresses the degree of grammar sharing across languages, as well as within language families. He highlights that the mere existence of an abstract syntax implies grammar sharing and he shows that some linguistic phenomena can be treated in a systematic way.

Vergne (2002) does not adhere to a grammar theory, but tries to reach language-independence by using an extremely **simple, minimalistic and radical approach** to building multilingual chunkers and (partial) parsers, without using full dictionaries. He shows the feasibility of his approach by building a tool that extracts subject-verb combinations for five languages, using dictionaries of only about 200 elements per language, case information and regular expressions matching certain combinations of word endings. More recently, Vergne (2009) proposed a chunker using only string length and word frequency, and applied it to 23 languages. The basic idea, which we share, is thus to limit the used resources to a bare minimum, i.e. to those elements that are required for a specific task.

It goes without saying that simple applications can more easily be achieved with simple means and that more complex applications are likely to benefit from a deeper linguistic analysis. There is thus not one solution for all tools and applications. However, we observed – for the information extraction tasks we are targeting – that even simple means can take you relatively far, and that minimalism and simplicity paid off for us.

**Related work: Machine Learning**

Machine Learning (ML) approaches have become very popular. Helped by the availability of more data and increased computer processing power, the technology has advanced a lot over the last years and the trend is likely to continue. The obvious appeal of self-learning software is that it will by itself take care of learning rules and vocabulary, and that it can be optimised for real-life data by training it on such data. ML is thus a very promising solution to achieve high multilinguality.

In the field of Machine Translation (MT), statistical (i.e. self-learning) methods are currently the major paradigm, i.e. systems that learn automatically from texts that have previously been translated manually. *Google translate*[11] now offers all language pair combinations for the impressive number of 58 languages, i.e. 1653 language pairs (status: February 2011). Never before has any translation software been available for so many languages. A current trend is to combine purely statistical MT with symbolic MT, e.g. by integrating the processing of syntactic rules (e.g. Goutte et al. 2009). When doing this, the question arises again how this can be done with minimal effort for many languages, but presumably the rules will be rather language or language pair-dependent.

In the field of Named Entity Recognition (NER), ML techniques have been widely used (Nadeau & Sekine 2009). The most common approach is to use *supervised ML*, i.e. training a system on previously annotated corpora. While the idea is attractive, the de-facto limitation is the fact that producing such annotated corpora (e.g. for new languages) is labour-intensive and expensive. Alternatives are to use semi-supervised or unsupervised learning methods. *Semi-supervised* learning involves a set of seeds to start the learning process and boot-strapping methods to gradually increase the number of patterns and resources. *Unsupervised* learning makes use of external resources and observations, and especially of large corpora. An example for such work applied to NER is that of Shinyama & Sekine (2004), who tried to detect named entities based on the observation that a named entity is likely to appear synchronously in several news articles, whereas common nouns have different distribution patterns. An open issue is how to combine ML methods with manual intervention, e.g. if one wants to manually correct and improve the machine-learnt recognition rules.

---

[11] http://translate.google.com/

ML methods, especially semi-supervised and unsupervised, are clearly very promising when attempting to achieve high multilinguality. In the context of EMM, however, we decided for ourselves to use hand-crafted rules, and to enhance manually produced dictionaries and word lists by using bootstrapping and Machine Learning methods. Doing this allows us to keep control over the recognition performance. Most recent publications on IE describe ML methods, often highlighting the language-independence of the described approach. However, through personal communication with many system developers, we got the impression that most existing multilingual IE applications are in fact rule-based, or – like in EMM – that the use of ML is restricted to lexical acquisition.

We believe that our approach requires less time per language than when using pure Machine Learning methods. We typically invest a maximum of three person months to add a new language to the tool set, as this is the average time of having a native speaker trainee available to us. In this time period, the person can discover and add news sources, translate the Boolean category definitions used in EMM-NewsBrief and in MedISys, provide the linguistic IE resources for the new language, and test the performance. However, it is also possible to produce reasonable initial linguistic resources to recognise named entities and quotations in a new language within one working week. Information redundancy is high in EMM, so that we aim at high precision and accept lower recall, assuming that, if we miss some information in one article, we are likely to find it in another.

**Related work: Cross-lingual projection**

The shortage of annotated multilingual data that can be used to train or evaluate IE tools in various languages is sometimes addressed through cross-lingual projection, using parallel corpora. The idea is to benefit from the availability of data in a resource-rich language such as English and to project the English annotations into the other language(s). For instance, Bentivogli et al. (2004) project semantic word sense annotations from English to Italian, using a bilingual parallel text collection and word-alignment tools. Ehrmann & Turchi (2010) aim to overcome the shortage in annotated NE data by projecting NE annotations from English documents to French, German, Spanish and Czech documents, using a multi-parallel corpus and word-aligning the languages with a phrase-based statistical Machine Translation system. Turchi et al. (2010), finally, create a seven-language gold standard document collection to evaluate multilingual multi-document extractive summarisation software, by manually selecting the most important English sentences from each cluster and by projecting the selection into the sentence-aligned target language documents. When using this gold standard collection to evaluate their multilingual summariser, they made an interesting observation: Their purely statistical – and thus in principle language-independent – tool performed rather differently across languages, which is unexpected as the general assumption would be that the performance should be comparable. This insight would not have been possible without using a parallel document collection allowing the accurate comparison of text mining results across languages.

Having access to multilingual gold standard data is obviously very important in the highly multilingual EMM setting. Annotation projection is an obvious and promising way of generating such evaluation data. The biggest bottleneck is the lack of parallel corpora covering more than only two or three languages.

**Insights by EMM developers**

Due to the strict requirement of having to analyse documents in many languages (ideally, all 23 official EU languages, plus more) while working in a small team (three computational linguists during most of the years, but currently seven), we always had to use minimalistic methods and try to achieve with them as much as possible. Basically, we were reduced to *not* using parsers, part-of-speech taggers, morphological analysers and full dictionaries for any of the languages, and we had to keep the effort of adding a new language to the tool set to about three months, including testing. While good linguistic resources are available freely for some languages, we could not make use of them as we needed to keep the work parallel for all languages. The kind of resources we *do use* are targeted word lists (name titles; gazetteers of place names; sentiment words; reporting verbs and – very important – different types of stop words, etc.); mixed-language Boolean combinations of category-defining words; the output of our own NER tools; statistics, heuristics, boot-strapping methods and machine learning.

Regarding methods to keep the development effort per language down, we basically had the same insights other groups identified (i.e. those mentioned in the first paragraph in section 0). The most important ones for us are *modularity* and *simplicity*. Another principle we often applied, closely linked to simplicity, is *under-specification*. The idea is: don't formulate constraints if you don't urgently need them, as they are time-consuming to produce and they may hinder you in your analysis of other languages. For instance, if it is not strictly necessary in local patterns to specify the morphological agreement and the order of words or word groups (e.g. modifiers for titles in person name recognition), simply leave them unspecified (see also Section **Error! Reference source not found.**).

Another difference to the work presented in 0 is that we developed further the idea of using mostly *language-independent* rules that make reference to *language-specific* resource files containing application-focused word lists. For applications such as person and organisation name recognition, quotation recognition, and for geo-tagging and grounding (distinguishing, e.g., which of the 15 locations world-wide with the name of *Paris* is being referred to in the text), this principle was adhered to quite closely. In exceptional cases, such as person name recognition in Arabic (which does not distinguish upper and lower case), separate recognition patterns were added and located in the file containing the language-specific information (Zaghouani et al. 2010). That way, the resulting system is entirely modular. When adding a new language, it is normally sufficient to plug in the language-specific parameter file. For person name recognition, this file includes long lists of words, phrases and regular expressions that are typically found next to person names and that help determine whether some uppercase words are a name or not. The resulting patterns can also identify and store names and titles in more complex expressions such as: *the recently elected chairperson of LREC, Nicoletta Calzolari*, or *Tony Blair, 57-year old former British Prime Minister*. The required word lists are usually produced using seed patterns, machine learning and knowledge discovery, and boot-strapping, but external knowledge sources such as Wikipedia are of course also used, when available.

Highly inflected languages are a challenge for simple methods that rely a lot on matching expressions in a text against word lists. To solve the problem, we either apply some simple language-dependent suffix stripping and suffix replacement rules (e.g. to recognise *New Yorgile* as an Estonian inflection of the name *New York*), or we pre-generate many variants of known names so as to facilitate their recognition in text, using finite state tools. Our data base contains over 1 million known entities (plus additional hundreds of thousands of known name variants), collected through multi-annual multilingual information extraction. For example, for the name part (Tony) *Blair* and the Slovenian language, inflections such as the following are automatically generated: *Blairom, Blairju, Blairjem*, etc.

For the more complex task of event scenario template filling in seven languages (Arabic, English, French, Italian, Portuguese, Russian and Spanish), we did not entirely adhere to the principle of language-independent grammars (Tanev et al. 2009). However, the approach still is minimalistic in the sense that no part-of-speech taggers or syntactic parsers are used and that we do not use complete dictionaries. Instead, the system uses local grammars to identify the information for the individual slots, such as: event type; number, status and type of victims; perpetrator; weapon; location and time. This information is then combined to produce the entire event description.[12]

The approach for the development of multilingual text mining applications in EMM is described in more detail in Steinberger et al. (2008), where we also give an overview of how these generic principles work in practice, for seven different text mining applications. In Steinberger et al. (forthcoming), we describe the concrete effort of adding a new language to the tool set: the African Bantu language Swahili.

EMM-NewsExplorer also offers some cross-lingual functionality for its twenty languages, i.e. cross-lingual cluster linking, name variant matching (including across scripts), and merging the information extracted about entities in all monitored languages. As there are 190 language pairs for 20 languages, the use of bilingual resources and methods needed to be strictly avoided. Another guideline we follow is thus: for cross-lingual applications, avoid the usage of bilingual resources and favour (more or less) language pair-independent methods (see also Section **5.1**).

It should be clear by now that EMM tools do not adhere to a grammar theory or any other theoretical framework.

# 5. Examples for applications developed according to these guidelines

The means imposed by the multilinguality requirement, presented in Section 0, are very restrictive. While they make extending to many languages easier, they also represent a challenge for most text mining applications. In the previous section, it already became clear how we solved the challenge for person name recognition and event scenario filling. We will now try to sketch solutions for two application we have developed already (name variant matching and quotation recognition;

---

[12] The event extraction results are accessible at http://emm.newsbrief.eu/geo?type=event&format=html&language=all .

Sections **Error! Reference source not found.** and **XXX**), and for others we are currently working on (Sentiment Analysis, 0; and Multi-document Summarisation, 0).

**Matching name variants across many languages and scripts**

The NER tool described in Section 0 recognises names in currently 20 languages. It happens frequently that names for the same person are spelled differently, not only across scripts (Arabic, Cyrillic, Roman) and languages, but even within the same language. Figure 2 (**XXX**) shows some of the many spelling variants for the same entity (in the section *Names*). As the aim is to index documents by the entities mentioned and to establish links between entities independently of their spelling, it is important to identify that all of these spellings are simply variants of the same name. The challenge thus is to detect automatically that the names (Nikita) *Krushchev*, *Chruschtschow*, *Chrusjtjov*, *Hruščov* and many more are all name variants of the name of the former Soviet leader

Никита Хрущев.[13] Establishing which spelling variants belong to the same name is typically done through machine learning: based on bilingual lists of names and their translation or transliteration, software learns equivalences of characters and character groups (e.g. Lee et al. 2006). This approach works well, but its restrictions are that it requires long parallel lists of names for training and that the learning is language pair-specific, making it difficult to deal with 20 languages and 190 language pairs. In EMM-NewsExplorer, where we need to decide which of the hundreds of newly found names every day may be variants of any one of the over one million known names and name variants in the EMM name database, we solved the problem in a multi-step process, which is the same for all input languages (see **Figure 3**): (1) If the name is not written using the Roman script: Transliteration into the Roman script (using standard n-to-n character transliteration rules); (2) name normalisation; (3) vowel removal to create a consonant signature; (4) for all names with the same consonant signature, calculate the overall similarity between each pair of names, based on the edit distance of two representations of both names: between the output of steps of (1) and (2). If the overall similarity of two names is above the empirically defined threshold of 0.94, the two names are automatically merged. If the similarity lies below that value, they are kept as separate entities. The normalisation rules (see **Figure 4**) are hand-drafted, based on the observation of regular name spelling variations. The method for normalisation and variant mapping is the same for all languages and all rules apply to all languages. For details on this name variant matching process and a list of reasons for the existence of name variants, see Steinberger & Pouliquen (2007).

Latin normalisation:
- accented character → non-accented equivalent
- double consonant → single consonant
- ou → u
- " al-" →
- wl (beginning of name) → vl
- ow (end of name) → ov
- ck → k
- ph → f
- ž → j
- š → sh
- x → ks

Remove vowels

Malik al-Saïdoullaïev
Malik al-Saidoullaiev
Malik al-Saidoulaiev
Malik al-Saidulaiev
Malik Saidulaiev
... mlk sdlv

| Name | Normalised form |
| --- | --- |
| Mohammed Siad Barre, Mohamed Siad Barré, Мохаммед Сиад Барре, محمد سياد بري | **mhmd sd br** (mohamed siad bare) |
| Mahmoud Ahmadinejad, Mahmūd Ahmadīnežād | **mhmd hmdnjd** (mahmud ahmadinejad) |

---

[13] See the NewsExplorer entity page http://emm.newsexplorer.eu/NewsExplorer/entities/en/7472.html .

**Figure 4.** Selection of name normalisation rules and their result. The hand-crafted rules are based on empirical observations about regular spelling variations. They are purely pragmatically motivated and not intended to represent any linguistic reality.

**Quotation Recognition**

The quotation recognition tool, covering 20 languages, aims to detect occurrences of direct reported speech if the speaker can be unambiguously identified (for display in *NewsBrief* and on the person pages in *NewsExplorer*[14]). If the quotation makes reference to another known entity, this will be recorded, as well (quotation *about* an entity). Details on this tool can be found in Pouliquen et al. (2007b). The patterns make reference to quotation markers (e.g. ", ', «), person or organisation names identified in the same article, reporting verbs (e.g. *said, reported, argues*, etc.) and a range of modifiers that can be found between any of the other elements (e.g. *yesterday*, *on TV*, etc.). The simplified sample rule below would successfully identify the quotation, the speaker (Angela Merkel) and the entity referred to in the quotation (Barack Obama) in the following string: *Merkel said yesterday on TV "…Obama …"*.

NAME    REPORTING_VERB    MODIFIER    "QUOTE"

Note that the co-reference between *the US President* or *President Obama* and the known entity *Barack Obama* will be established if the full name is mentioned at least once in the document and if either at least one name part and/or one of the many previously identified titles for that name are found.

To comply with the *simplicity* and *under-specification* requirement, the order of modifiers and any morphological agreement (e.g. in number or gender) will not be specified. It is furthermore possible to allow any combination of individual modifier words (e.g. *TV yesterday on*) without much risk as we focus on recognition (and not generation) and the ungrammatical combinations will simply not be found in real-life text.

**Sentiment analysis**

EMM users are not only interested in factual content, but also in opinions on certain entities and issues (such as the EU constitution). Questions asked concern the (positive or negative) attitude of media sources in certain countries towards these *targets*, and of changes across languages and over time. Approaches to opinion mining vary widely regarding the methods and the depth of analysis (see, e.g. Pang & Lee 2008). Due to our multilinguality requirement, we again need to use the simplest possible methods, involving the usage of word lists (positive and negative words, polarity inverters, strength enhancers and diminishers) and previously recognised named entities). To avoid negative news content (e.g. in news on natural disasters) having an impact on the detected sentiment towards any entity mentioned in these news items, we decided not to consider sentiment words that are also part of EMM's category-defining terms, such as *disaster, tsunami* and *flood* for the EMM category 'Natural Disasters'. These category-defining terms are not ideal for the task of distinguishing good or bad news content from positive or negative sentiment, but they are readily available for all EMM languages. To ensure furthermore that the sentiment words actually apply to the entity we are interested in, we use word windows around the entities and their titles. Experiments with various English language sentiment vocabularies showed that the best-performing results were achieved with a window size of six words to either side of the entity and its titles. See Balahur et al. (2010) for details.

---

[14] See, for example, Barack Obama's page at  http://emm.newsexplorer.eu/NewsExplorer/entities/en/1510.html .

Many English language sentiment dictionaries are freely available, but such vocabulary lists are scarce for other languages. Having identified a reasonably performing language-independent method for sentiment analysis, we are currently working on semi-automatically generating large non-English sentiment vocabularies.

**Multilingual multi-document summarisation**

Due to the high redundancy of EMM's news content (100,000 news articles per day collected from about 2,500 different media sources), a major task performed by the EMM systems is to group related articles into clusters, and to track the development of these news clusters over time (*topic detection and tracking*). Currently, EMM displays the title and description of each cluster's centroid article, but a proper summary per cluster, and update summaries for clusters related over time, would be very useful. This was the motivation to work on multilingual multi-document summarisation.

As abstractive summarisation would require many linguistic resources, our multilingual environment restricts us to extractive methods, not considering syntax. The proposed solution consists of using latent semantic analysis (LSA) to select the most informative sentences from the whole cluster (similar to Gong & Liu 2002). To reduce redundancy in the summaries, the information covered in already selected sentences is subtracted from the LSA vector representation in order to ensure that the next sentences contain new information. The approach is thus based on a language-independent vector representation. However, in addition to a list of words and word-ngrams per sentence, the LSA input in our system consists of previously identified entity mentions, and of (non-disambiguated) mentions of terms from the multilingual MeSH thesaurus (*Medical Subject Headings*[15]). The idea behind this approach is (a) to give higher weight to entities and (b) to capture some synonymy and hyponymy relations, both to select the most important sentences and to avoid information redundancy in the selected sentences. Due to our historical collection of multilingual name variants and a list of previously found titles for each entity, our lookup recognises name mentions even if the spelling varies. The approach was successful at the TAC'2009 competition, achieving second place in the most important category *overall responsiveness*, out of 54 submissions. For an overview of that system, see Kabadjov et al. (2010).

## 6. Required language resources

In the previous sections, we tried to summarise the constraints we imposed on ourselves when developing multilingual text mining applications. We also tried to sketch simple solutions that allowed us to avoid using too many linguistic resources. If linguistic resources had been freely available for all the languages we are trying to cover, development time would have been reduced and it is likely that the results achieved would be better. In this section we thus want to give an idea of tools and resources that – we believe – would enable the community to build multilingual text mining applications better and more quickly.

The major – probably banal – statement we would like to make is that the community would strongly benefit from *freely available, simple, parallel and uniform multilingual dictionaries, corpora and software tools*.

The resources should ideally be *free* because universities and research organisations in many countries would otherwise not get access to these resources. This is particularly true for lesser-used languages, which are the majority of languages. The current situation leads to a scientific brain drain because students and researchers around the world have to work on (mostly) English language applications because this is one of the very few languages for which tools are readily available. If working on their own languages, they would be reduced to developing basic tools and resources such as corpora, dictionaries and morphological analysers.

---

[15] See http://www.nlm.nih.gov/mesh/. The multilingual MeSH term recognition software was developed by Health-on-the-Net (HON, http://www.hon.ch/).

The tools and resources should be *simple* because they would otherwise never be built for many languages. We believe this to be true because of the associated cost, the time required for the development, and the limitations on available qualified manpower. At a recent FLaReNet event[16], Grefenstette (2010) presented the idea of a community-based Web 2.0 effort to build simple dictionaries for many languages. The basic idea is to ask native speakers to provide lemma, main part(s)-of-speech and English translation(s) for a list of (possibly frequency-sorted) word surface forms. The usual Web 2.0 incentives and control mechanisms could be applied and the resource could be downloadable anytime by anyone. Even non-linguists can provide this type of information. Usability would be limited for more complex applications requiring, for instance, sub-categorisation frames, but applications like those developed as part of EMM would certainly benefit. Grefenstette's pragmatic proposal of also providing the English translation is probably the most arguable feature. Amendments to his ideas may also be useful for compounding and agglutinative languages.

The tools and resources should be *parallel and uniform*, i.e. input and output format should be the same for all languages, the same set of parts-of-speech and syntactic categories should be used for all, etc. Ideally, resources should also be linked across languages. Uniform and parallel dictionaries would allow, for instance, writing multilingual rules and patterns much more easily. Successful efforts that produced such lexical resources in the past were Multext[17], Multext-East[18], GeoNames[19] and the various multilingual WordNet-related projects[20]. The *Eurovoc thesaurus*[21], a multilingual categorisation scheme with over 6,000 classes used by parliaments in Europe, was not developed for machine use, but it is still very useful because it covers almost thirty languages and it has been used to manually classify large numbers of documents. Using such uniform lexical resources, multilingual grammars are likely to be much more comparable and the effort of adapting a grammar to another language would be minimised.

---

[16] See http://www.flarenet.eu/?q=node/347.
[17] http://www.issco.unige.ch/en/research/projects/MULTEXT.html
[18] See http://nl.ijs.si/ME/
[19] See http://www.geonames.org/
[20] See http://www.globalwordnet.org/
[21] See http://europa.eu/eurovoc/ .

Parallel corpora are also much more useful than multi-monolingual corpora. Apart from their usefulness to train statistical machine translation and to construct multilingual dictionaries, they can be exploited to train and evaluate systems for information extraction, alignment, document categorisation, and more, with minimal effort. In spite of its limited subject domain, the 22-language parallel corpus *JRC-Acquis* (Steinberger et al. 2006) has therefore been useful for various multilingual tasks. As shown by Ehrmann & Turchi (2010) for Named Entity Recognition and by Turchi et al. (2010) for multi-document summarisation, annotations in one language version of a parallel corpus can be projected to the other languages, thus considerably saving annotation effort and creating a multilingual parallel training and evaluation resource. When evaluating any text mining tool on such a parallel resource, the performance across languages can be compared directly and fairly because the otherwise unknown parameters corpus size, text type, varying frequency of linguistic phenomena, etc. are the same for all languages. The tests carried out by Turchi et al. (2010), for instance, showed that the purely statistical software for multilingual multi-document summarisation produced rather different results for different languages on such a parallel corpus, raising questions regarding the common universality assumptions of language-independent software.

In the CoNLL shared tasks 2006 and 2007 (Nivre et al. 2007), dependency parsers were trained and tested for 13 and 10 languages, respectively. This was a very useful effort for creating resources, promoting multilinguality, and more. However, as the training corpora used different grammatical features and labels (e.g. for part-of-speech and syntactic phrases), the output for the same parsing system is not homogeneous across languages. Any rules reading the dependency tree output would thus need to be written differently for each language. This limits the usability of the otherwise very useful multilingual tool enormously. Software tools trained or built with uniform and parallel resources are likely to be parallel, or at least very similar, themselves. They would minimise any effort of building upon their output considerably.

It is also important to have a *single access point for licensing* issues (such as ELDA[22] and LDC[23]) to avoid having to contact many different content providers when building a highly multilingual system, although the usage entirely without licences would, of course, allow even more flexibility. Last, but not least, continuity of secure funding is obviously an important development factor for highly multilingual applications: Universities and other organisations receiving project-specific funding do not usually have the opportunity to extend their work to larger numbers of languages as they keep having to work on new areas.

---

[22] See http://www.elda.org/
[23] See http://www.ldc.upenn.edu/

It goes without saying that building resources and tools with these specifications is expensive and time-consuming. The number of highly multilingual parallel texts is limited and copyright issues may make it difficult to use them. The existence of the resources and tools described here may remain a dream. However, we feel that such resources would be a big step towards developing highly multilingual text mining applications, and awareness may be the first step towards achieving this goal.

There has been a lot of progress recently in the field of multilinguality and multilingual resources, which gives us hope that – also from a linguistic point of view – this world will soon be much smaller. Past and present initiatives such as FLaReNet[24], CLARIN[25], CLEF[26], ENABLER[27], META-Net's resource initiative META-SHARE[28], LDC's *Less Commonly Taught Languages* project[29] and the *Global WordNet Association*[30] are very promising and encouraging.

## 7. Summary and Conclusion

We have tried to show that there is a strong need for highly multilingual text mining applications (10, 20 or more languages), but that most available and operational systems cover only one or a small number of languages. Assuming that this is mostly due to the fact that the development of natural language processing tools for each language is time-consuming and expensive, we asked the question how the development effort per language can be minimised. The major tips and ideas we found in publications and personal discussions with multilingual system developers are: (a) keep your system modular; (b) keep the system simple, not only from a user's point of view, but also from that of the developer; (c) try to use uniform input and output structures; (d) use shared token classes, ideally based on surface-oriented features; (e) try to share grammar rules and lexical resources between languages; and (f) try to be minimalistic by providing and using only the type of information really needed for the application, rather than filling the whole paradigm (e.g. use partial dictionaries rather than trying to produce a complete lexicon for a language). Several developers of multilingual parsers furthermore pointed out the advantage of (g) adhering to grammar theories, as these allow stipulating general principles that apply to whole groups of languages, i.e. another type of grammar sharing. From an architectural point of view, however, the point was made that a theory-neutral approach is more flexible and lends itself more to a reuse of resources. While developing various text mining tools in up to twenty languages for the *Europe Media Monitor* (EMM) family of applications, we furthermore got convinced that it is useful and efficient (h) to write language-independent rules that make use of information stored in language-specific parameter files; (i) to under-specify wherever possible, in order to save time and not to use restrictions that may get in the way when dealing with another language.

---

[24] See http://www.flarenet.eu/
[25] See http://www.clarin.eu/
[26] See http://www.clef-campaign.org/
[27] See http://www.enabler-network.org/
[28] See http://www.meta-net.eu/meta-share
[29] See http://projects.ldc.upenn.edu/LCTL/
[30] See http://www.globalwordnet.org/

In the case of EMM tools, these requirements basically mean that the use of language-specific linguistic resources and tools should be minimised. We thus limited ourselves to work with restricted word lists, lookup procedures, machine learning and bootstrapping methods. Such simple means are rather restrictive and challenging. To show what can and what cannot be done adhering to these restrictions, we sketched the solutions adopted in a few of our own multilingual text mining applications.

We saw that machine learning solutions are particularly promising to achieve high multilinguality, but that the need for pre-tagged training data limits at least supervised learning methods to those few languages for which tagged corpora are available. Semi-supervised or unsupervised methods are, in principle, better suited for lesser-used languages, for which few linguistic resources exist. As the output of automatically learnt classifiers and rules cannot normally be easily improved and amended, we suggested the hybrid solution of using hand-crafted rules and to use Machine Learning to acquire the lexical entries.

We finally presented our own – probably unrealistic – opinion regarding the types of linguistic resources that would be useful to allow the computational linguistics community to develop more highly multilingual text mining applications more quickly, and why. These resources can be described as *freely available, simple, parallel and uniform multilingual dictionaries, corpora and software tools*. The number of current efforts and projects to produce multilingual resources shows a positive and encouraging trend.

There is more than one possible solution to overcome the multilinguality barrier, and each application has its own specific requirements. We hope, though, that this collection and discussion of ideas and insights may be useful for multilingual system developers.

## Acknowledgements

I would like to thank the following persons for having shared their own multilingual grammar writing experience with us, or their views on linguistic resources: Kalina Bontcheva (Sheffield University) on *GATE;* Frédérique Segond, Caroline Hagège and Claude Roux (Xerox Research Centre Europe) on the *Xerox Incremental Parser*; Aarne Ranta (Gothenburg University) on the *Grammatical Framework*; Jacques Vergne (Caen University) on sentence chunking using extremely light-weight methods; Eric Wehrli (Geneva University) on his deep-linguistic parser; Gregory Grefenstette (Exalead) and Gregor Thurmair (Linguatec) on their respective multilingual products; Khalid Choukri (ELRA/ELDA) and Gregory Grefenstette on linguistic resources; and my JRC colleagues Maud Ehrmann, Vanni Zavarella and Hristo Tanev for sharing their experiences and for their feedback on earlier versions of the paper. The ultimate responsibility for any errors, however, lies with me.

I would furthermore like to thank my superiors Erik van der Goot and Delilah Al Khudhairy for their support, and my colleagues in the OPTIMA group at the JRC for the fruitful and efficient collaboration over the past years, and for so reliably providing large amounts of clean multilingual news data, which allowed us to run many multilingual experiments. Building the complex EMM applications was a successful team effort that also includes many less rewarding and less visible tasks. My specific thanks go to my former colleague Bruno Pouliquen (now at WIPO in Geneva). We developed most ideas together, and he very efficiently implemented many ideas and integrated the many tools with each other.